\documentclass[conference]{IEEEtran}
\IEEEoverridecommandlockouts
\usepackage{cite}
\usepackage{amsmath,amssymb,amsfonts}
\usepackage{algorithmic}
\usepackage{algorithm}
\usepackage{graphicx}
\usepackage{textcomp}
\usepackage{xcolor}
\def\BibTeX{{\rm B\kern-.05em{\sc i\kern-.025em b}\kern-.08em
		T\kern-.1667em\lower.7ex\hbox{E}\kern-.125emX}}

\makeatletter
\let\old@ps@headings\ps@headings
\let\old@ps@IEEEtitlepagestyle\ps@IEEEtitlepagestyle
\def\confheader#1{%
	\def\ps@IEEEtitlepagestyle{
		\old@ps@IEEEtitlepagestyle
		\def\@oddhead{\strut\hfill#1\hfill\strut}
		\def\@evenhead{\strut\hfill#1\hfill\strut}
	}
	\ps@headings
}
\makeatother
\confheader{
	\small{Proceedings of the 13th RSI International Conference on Robotics and Mechatronics (ICRoM 2025), Dec. 16-18, 2025, Tehran, Iran} 
}
\usepackage[pscoord]{eso-pic}
\newcommand{\placetextbox}[3]{
	\setbox0=\hbox{#3}
	\AddToShipoutPictureFG*{ \put(\LenToUnit{#1\paperwidth},\LenToUnit{#2\paperheight}){\vtop{{\null}\makebox[0pt][c]{#3}}}
	}
}
\placetextbox{.5}{0.055}{\textbf{\small{Proceedings of the 13th RSI International Conference on Robotics and Mechatronics (ICRoM 2025), Dec. 16-18, 2025, Tehran, Iran}}}

\begin{document}
\title{Dynamic Modeling and LQR Control of a Single-Coaxial Drone with 2-DOF Thrust Vectoring Mechanism \\}

\author{\IEEEauthorblockN{Ali Jokar}
	\IEEEauthorblockA{\textit{Sharif AgRoLab} \\
		\textit{School of Mechanical Engineering} \\
		\textit{Sharif University of Technology}\\
		Tehran, Iran \\
		ali.jokar@sharif.edu}
	\and
	\IEEEauthorblockN{Amin Talaeizadeh}
	\IEEEauthorblockA{\textit{Sharif AgRoLab} \\ 
		\textit{School of Mechanical Engineering} \\
		\textit{Sharif University of Technology}\\
		Tehran, Iran \\
		amin.talaeizadeh@sharif.edu}
	\and
	\IEEEauthorblockN{Aria Alasty}
	\IEEEauthorblockA{\textit{Sharif AgRoLab} \\ 
		\textit{School of Mechanical Engineering} \\
		\textit{Sharif University of Technology}\\
		Tehran, Iran \\
		aalasti@sharif.edu}
	}
	
\maketitle

\begin{abstract}
Coaxial rotor drones have generated considerable interest because of energy efficiency and small size, but they are afflicted with inherent underactuation for roll and pitch control, although systems like swashplates have circumvented this limitation at the cost of greater mechanical complexity. This work presents a novel coaxial drone supplemented by a two-degrees-of-freedom pendulum mechanism for active thrust vectoring that offers a less mechanically complicated alternative. We develop a comprehensive Lagrangian dynamic model that does not ignore the inertial contributions of all the components, including body, servo arms, and motor assembly. A Linear Quadratic Regulator (LQR) is designed based on the linearized dynamics around the hover equilibrium. High-fidelity simulations taking actuator dynamics and sensor noise into account validate the proposed architecture. An Extended Kalman Filter (EKF) blends GPS, barometer, and IMU estimates with high accuracy for state estimation. The findings verify the potential and reliability of this approach for power-saving, rapid coaxial UAVs.
\end{abstract}

\begin{IEEEkeywords}
Coaxial Drone, LQR Controller, 2-DOF Pendulum Mechanism, EKF 
\end{IEEEkeywords}

\section{Introduction}
Coaxial rotor drones gained significant attention because of their mechanical complexity, compact size, and energy efficiency compared to quadcopters \cite{b1,b2}. However, their intrinsic underactuation—no roll and pitch control independently—has long limited their maneuverability and practical applications. Overcoming this intrinsic drawback, several solutions have been proposed by researchers including auxiliary control surfaces \cite{b3}, tilting rotors \cite{b4,b5}, and swashplate mechanisms. Such approaches have yielded mixed success in terms of attitude control, but at the common cost of increased mechanical complexity or reduced efficiency.

Other somewhat more recent efforts have been directed at coaxial configurations with active thrust-vectoring capability. North \cite{b6} documented a coaxial UAV model with hardware realization but with linearized simplified dynamics and with no allowance for the complete inertial effects of translating parts. Chen et al. \cite{b2} designed a coaxial quadcopter using two servo motors to tilt the thrust axis and control pitch and roll. Although they derived the dynamics and experimentally verified their design by simulations and flight tests, the model neglected inertial contributions of the motor assembly and servos, which are generally large for aggressive maneuvers.

Other pertinent works include Glida et al. \cite{b7}, who developed a terminal sliding mode controller without a model for a coaxial UAV; Wei et al. \cite{b8}, who simulated a hybrid PID-sliding mode controller on a prototype; and Bernardes et al. \cite{b9}, who simulated a single-rotor UAV with a swashplate. Besides, coaxial rotor aerodynamics articles \cite{b10},\cite{b11,b12} and underactuated systems control articles \cite{b13},\cite{b14,b15} are basic works.

Despite these advances, a full dynamic model of inertial coupling of all flight masses in a coaxial thrust-tilting UAV remains yet unknown. The paper fills the gap by:
\begin{itemize}
	\item Establishing a full Lagrangian model of body masses and inertias, both the servo arms, and motor-prop assembly.
	\item Deriving an LQR controller from the linearized model around hover.
	\item High-fidelity validation of simulation system with actuator dynamics, ground contact, and sensor noise utilizing EKF-based state estimation.
\end{itemize}

The main contribution of this work is the use of a two-degrees-of-freedom pendulum mechanism for thrust vectoring, which is mechanically simpler and potentially more robust than the conventional swashplate system but with identical attitude control authority.

The paper has been organized in the following manner: Section II addresses dynamic modeling of the UAV by Lagrangian formulation. Section III presents controller design through linearization and LQR. Section IV presents simulation results with system performance. Section V finally reports the findings and concludes with suggestions on future work.

\section{Dynamic Modeling via Lagrangian Formulation}
\begin{figure}[htbp]
	\centerline{\includegraphics[width=7cm,height=8cm]{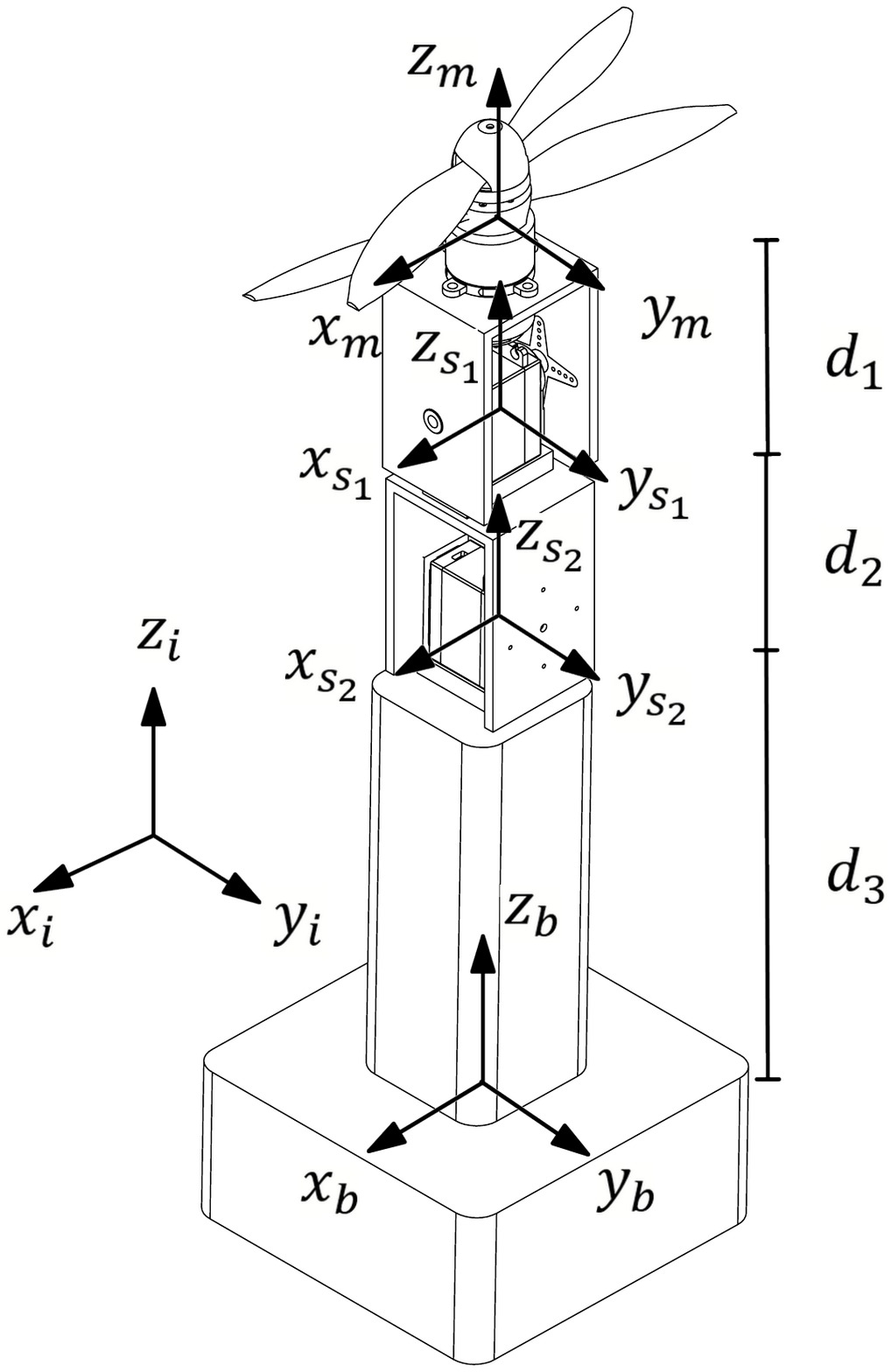}}
	\centering
	\caption{Schematic of a coaxial drone with coordinates indicating the pendulum mechanism structure}
	\label{quadcopter_schematic}
\end{figure}

Figure \ref{quadcopter_schematic} illustrates the mechanical architecture of the proposed coaxial UAV, detailing the main body, servo arms, and motor-propeller assembly. The core innovation is a two-degree-of-freedom pendulum mechanism that enables thrust vectoring. This is achieved by actuating servo motor $S_2$ to adjust the pitch angle $\theta_p$ (rotation about the $y_{s_2}$ axis) and servo motor $S_1$ to adjust the roll angle $\theta_r$ (rotation about the $x_{s_1}$ axis). This coordinated tilting of the motor assembly provides a mechanically elegant solution to the roll and pitch control problem inherent in conventional coaxial designs.

The UAV is modeled as a four rigid-body system: body (of mass $m_b$), two servo arms (of masses $m_{s_2}$ and $m_{s_1}$), and motor-propeller assembly (of mass $m_m$). The generalized coordinates are:
\begin{equation}
	\mathbf{q} = [x, y, z, \phi, \theta, \psi]^T,
\end{equation}
where $(x,y,z)$ is the position of the main body center of mass, $(\phi,\theta,\psi)$ are Euler angles (roll, pitch, yaw) in the $z-y'-x''$ rotation sequence, and $\theta_p$, $\theta_r$ are the pitch and roll servo angles.

\subsection{Kinetic and Potential Energy}
The rotation about the local axes $z-y^{\prime}-x^{\prime\prime}$ included, the rotation matrix from the body frame to the inertial frame with the roll ($\phi$), pitch ($\theta$), and yaw ($\psi$) angles is given by Equation (2):
\begin{equation}
\begin{aligned}
		&\textbf{R}_b^i =R_x(\phi) R_y(\theta) R_z(\psi) = \\ &
	\begin{bmatrix}
		C\psi C\theta & C\psi S\phi S\theta - C\phi S\psi & S\psi S\phi + C\phi C\psi S\theta \\
		C\theta S\psi & C\phi C\psi + S\phi S\theta S\psi & C\phi S\psi S\theta - C\psi S\phi \\
		-S\theta & C\theta S\phi & C\theta C\phi
	\end{bmatrix},
\end{aligned}
\end{equation}
where $R_b^i$ is the rotation matrix of body-to-inertial frame. S and C stand for sine and cosine respectively in Equation (2) and the same notation will be employed throughout the text.

The rotation matrices about each axis used in change of secondary to primary frame are as follows:
\begin{align}
	R_x &= 
	\begin{bmatrix}
		1 & 0 & 0 \\
		0 & C\beta & -S\beta \\
		0 & S\beta & C\beta
	\end{bmatrix},
	& \\
	R_y &= 
	\begin{bmatrix}
		C\beta & 0 & S\beta \\
		0 & 1 & 0 \\
		-S\beta & 0 & C\beta
	\end{bmatrix},
	& \\
	R_z &= 
	\begin{bmatrix}
		C\beta & -S\beta & 0 \\
		S\beta & C\beta & 0 \\
		0 & 0 & 1
	\end{bmatrix}.
\end{align}

Total kinetic energy $T$ is the sum of translational and rotational energy of all elements. Consider position vectors:
\begin{align}
	&r_b = [x,y,z]^T, \\
	&r_{G_{s_2}} = r_b + R_b^i \begin{bmatrix} 0 \\ 0 \\ d_3 \end{bmatrix}, \\
	&r_{G_{s_1}} = r_b + R_b^i \left( \begin{bmatrix} 0 \\ 0 \\ d_3 \end{bmatrix} + R_y^{\theta_p} \begin{bmatrix} 0 \\ 0 \\ d_2 \end{bmatrix} \right), \\
	&r_{G_m} = r_b + R_b^i \left( \begin{bmatrix} 0 \\ 0 \\ d_3 \end{bmatrix} + R_y^{\theta_p} \begin{bmatrix} 0 \\ 0 \\ d_2 \end{bmatrix} + R_y^{\theta_p} R_x^{\theta_r} \begin{bmatrix} 0 \\ 0 \\ d_1 \end{bmatrix} \right),
\end{align}
where $d_1,d_2,d_3$ are structural offsets. The angular velocities are:
\begin{equation}
	\begin{aligned}
		& \omega_b = [\dot{\phi}, \dot{\theta}, \dot{\psi}]^T, \\
	    & \omega_{s_2} = \omega_b, \\
		& \omega_{s_1} = \omega_b + [0, \dot{\theta}_p, 0]^T, \\
		& \omega_m = (R_y^{\theta_p})^T(\omega_b + [0, \dot{\theta}_p, 0]^T) + [\dot{\theta}_r, 0, 0]^T.
	\end{aligned}
\end{equation}

The total kinetic energy is:
\begin{equation}
	\begin{aligned}
		& T = \frac{1}{2} m_b \|\dot{r}_b\|^2 + \frac{1}{2} m_{s_2} \|\dot{r}_{G_{s_2}}\|^2 + \frac{1}{2} m_{s_1} \|\dot{r}_{G_{s_1}}\|^2 + \\ & \frac{1}{2} m_m \|\dot{r}_{G_m}\|^2 
		+ \frac{1}{2} \omega_b^T I_b \omega_b + \frac{1}{2} \omega_{s_1}^T I_{s_1} \omega_{s_1} + \\& \frac{1}{2} \omega_{s_2}^T I_{s_2} \omega_{s_2} + \frac{1}{2} \omega_m^T I_m \omega_m .
	\end{aligned}
\end{equation}

The potential energy is:
\begin{equation}
	\begin{aligned}
		& V = g ( m_b z + m_{s_2} (r_{G_{s_2}} \cdot \hat{z}_i) +\\ & m_{s_1}  (r_{G_{s_1}} \cdot \hat{z}_i) + m_m (r_{G_m} \cdot \hat{z}_i) ).
	\end{aligned}
\end{equation}

\subsection{Equations of Motion}
The Lagrangian is $L = T - V$. Equations of motion are obtained from:
\begin{equation}
	\frac{d}{dt} \left( \frac{\partial L}{\partial \dot{q}_i} \right) - \frac{\partial L}{\partial q_i} = Q_i,
\end{equation}
where $Q_i$ are generalized forces and moments computed by the principle of virtual work.

For the translational coordinates ($x, y, z$), the generalized forces are derived from the thrust vector projected into the inertial frame:
\begin{equation}
	Q_i = F \cdot \frac{\partial r_{G_m}}{\partial q_i},
\end{equation}
where $F$ is given by:
\begin{equation*}
	F = \begin{bmatrix}
		0 \\ 0 \\ k_f (\dot{{\theta}_1}^2 + \dot{{\theta}_2}^2)
		\end{bmatrix},
\end{equation*}
where $\dot{{\theta}_1}$ , $\dot{{\theta}_2}$ are angular velocity of propeller, and $k_f$ is the thrust coefficient.
The force $F$ must be expressed in the inertial frame as follows:
\begin{equation}
F^i = R_b^i R_y^{\theta_p} R_x^{\theta_r} F,
\end{equation}
Here, $q_i$ for $i = 1, 2, 3$ correspond to $x, y, z$.

For the rotational coordinates ($\phi, \theta, \psi$), the generalized moments are obtained by virtual work as:
\begin{equation}
Q_i = M^i_1 \cdot \frac{\partial \omega_{m_1}}{\partial \dot{q_i}} - M^i_2 \cdot \frac{\partial \omega_{m_2}}{\partial \dot{q_i}} + M^i_3 \cdot \frac{\partial \omega_{m}}{\partial \dot{q_i}},
\end{equation}
where $M_1$ and $M_2$ are given by:
\begin{equation}
M_1 = \begin{bmatrix}
0 \\ 0 \\ k_t \dot{{\theta}_1}^2
\end{bmatrix}, \quad
M_2 = \begin{bmatrix}
0 \\ 0 \\ k_t \dot{{\theta}_2}^2
\end{bmatrix},
\end{equation}
$k_t$ is the drag coefficient. $M_1$ and $M_2$ are expressed in the inertial frame as:
\begin{equation}
M^i_1 = R_b^i R_y^{\theta_p} R_x^{\theta_r} M_1, \quad
M^i_2 = R_b^i R_y^{\theta_p} R_x^{\theta_r} M_2,
\end{equation}
The term $M^i_3$ is computed as:
\begin{equation}
M^i_3 = R_b^i \left( 
\begin{bmatrix}
0 \\ 0 \\ d_3
\end{bmatrix} + R_y^{\theta_p} \begin{bmatrix}
0 \\ 0 \\ d_2
\end{bmatrix} + R_y^{\theta_p} R_x^{\theta_r} \begin{bmatrix}
0 \\ 0 \\ d_1
\end{bmatrix} \right) \times R_y^{\theta_p} R_x^{\theta_r} F,
\end{equation}
Here, $q_i$ for $i = 4, 5, 6$ correspond to $\phi, \theta, \psi$.

Applying the Lagrange formulation results in the following system:
\begin{equation}
	M(\mathbf{q})\ddot{\mathbf{q}} + C(\mathbf{q}, \mathbf{\dot{q}})\mathbf{\dot{q}} +G(\mathbf{q}) = Q,
\end{equation}
This system was implemented symbolically in MATLAB and converted to numerical functions.

\subsection{System Parameters}
The physical parameters of the system are shown in Tables \ref{tab:general_params}-\ref{tab:motor_specs}.The parameters include masses, inertias, geometric dimensions, and aerodynamic coefficients that characterize the UAV dynamics.
\begin{table}[H]
\centering
\caption{Physical and aerodynamic system parameters}
\begin{center}
\begin{tabular}{|c|c|c|}
\hline
\textbf{Parameter} & \textbf{Value} & \textbf{Unit} \\
\hline
Gravitational acceleration ($g$) & 9.81 & $\text{m/s}^2$ \\
Structural offset $d_1$ & 0.05 & m \\
Structural offset $d_2$ & 0.09 & m \\
Structural offset $d_3$ & 0.30 & m \\
Thrust coefficient ($k_f$) & $1.3 \times 10^{-3}$ & $\text{N} \cdot \text{s}^2$ \\
Torque coefficient ($k_t$) & $6.7 \times 10^{-4}$ & $\text{N} \cdot \text{m} \cdot \text{s}^2$ \\
\hline
\end{tabular}
\end{center}
\label{tab:general_params}
\end{table}
\begin{table}[H]
\centering
\caption{Main body physical specifications}
\begin{center}
\begin{tabular}{|c|c|c|}
\hline
\textbf{Parameter} & \textbf{Value} & \textbf{Unit} \\
\hline
Body mass ($m_b$) & 0.45 & kg \\
Moment of inertia $I_{xx}$ & 0.002 & $\text{kg} \cdot \text{m}^2$ \\
Moment of inertia $I_{yy}$ & 0.001 & $\text{kg} \cdot \text{m}^2$ \\
Moment of inertia $I_{zz}$ & 0.002 & $\text{kg} \cdot \text{m}^2$ \\
\hline
\end{tabular}
\end{center}
\label{tab:body_specs}
\end{table}
\begin{table}[H]
\centering
\caption{First servo arm physical specifications}
\begin{center}
\begin{tabular}{|c|c|c|}
\hline
\textbf{Parameter} & \textbf{Value} & \textbf{Unit} \\
\hline
Servo arm mass ($m_{s_1}$) & 0.07 & kg \\
Moment of inertia $I_{xx}$ & 0.00002 & $\text{kg} \cdot \text{m}^2$ \\
Moment of inertia $I_{yy}$ & 0.000025 & $\text{kg} \cdot \text{m}^2$ \\
Moment of inertia $I_{zz}$ & 0.00003 & $\text{kg} \cdot \text{m}^2$ \\
\hline
\end{tabular}
\end{center}
\label{tab:arm1_specs}
\end{table}
\begin{table}[H]
\centering
\caption{Second servo arm physical specifications}
\begin{center}
\begin{tabular}{|c|c|c|}
\hline
\textbf{Parameter} & \textbf{Value} & \textbf{Unit} \\
\hline
Servo arm mass ($m_{s_2}$) & 0.07 & kg \\
Moment of inertia $I_{xx}$ & 0.00002 & $\text{kg} \cdot \text{m}^2$ \\
Moment of inertia $I_{yy}$ & 0.000025 & $\text{kg} \cdot \text{m}^2$ \\
Moment of inertia $I_{zz}$ & 0.00003 & $\text{kg} \cdot \text{m}^2$ \\
\hline
\end{tabular}
\end{center}
\label{tab:arm2_specs}
\end{table}
\begin{table}[H]
\centering
\caption{Motor assembly physical specifications}
\begin{center}
\begin{tabular}{|c|c|c|}
\hline
\textbf{Parameter} & \textbf{Value} & \textbf{Unit} \\
\hline
Motor mass ($m_m$) & 0.16 & kg \\
Moment of inertia $I_{xx}$ & 0.00006 & $\text{kg} \cdot \text{m}^2$ \\
Moment of inertia $I_{yy}$ & 0.00004 & $\text{kg} \cdot \text{m}^2$ \\
Moment of inertia $I_{zz}$ & 0.00006 & $\text{kg} \cdot \text{m}^2$ \\
\hline
\end{tabular}
\end{center}
\label{tab:motor_specs}
\end{table}

\section{Controller Design: Linearization and LQR}
\subsection{Linearization}
Defining the state as ${X} = [\mathbf{q}^T, \dot{\mathbf{q}}^T]^T \in \mathbb{R}^{16}$ and control input as ${u} = [\tau_p, \tau_r, \dot{\theta}_1, \dot{\theta}_2]^T$, where $\tau_p$, $\tau_r$ are servo motor torques and $\dot{{\theta}_1}$, $\dot{{\theta}_2}$ are angular velocities of propellers.
The space model for equations as follows:
\begin{equation}
	\dot{{X}} = \mathbf{f}(X, u) = 
	\begin{bmatrix} 
\dot{\mathbf{q}} \\ {M}^{-1}({Q} - {C(\mathbf{q},\mathbf{\dot{q}})\mathbf{\dot{q}}} - G(\mathbf{q}))
\end{bmatrix}.
\end{equation}

The nonlinear dynamics are linearized around the hover equilibrium:
\begin{equation}
\mathbf{q_0} = [0,0,0,0,0,0]^T, \quad \dot{\mathbf{q_0}} = {0},
\end{equation}
with equilibrium inputs:
\begin{equation}
\tau_{p_0} = \tau_{r_0} = 0, 
\quad \dot{\theta}_{1_0} = \dot{\theta}_{2_0} = \sqrt{\frac{m_{\text{tot}} g}{2 k_f}},
\end{equation}
where $m_{\text{tot}} = m_b + m_{s_1} + m_{s_2} + m_m = 0.75~\text{kg}$. 
so the linearized model is:
\begin{equation}
\dot{X} = \mathbf{A} ({X} - {X}_0) + \mathbf{B} ({u} - {u}_0),
\end{equation}
where $\mathbf{A} = \frac{\partial \mathbf{f}}{\partial{X}}$ and $\mathbf{B} = \frac{\partial \mathbf{f}}{\partial {u}}$ are Jacobians evaluated at equilibrium.

\subsection{LQR Design}
The LQR controller minimizes:
\begin{equation}
J = \int_0^\infty \left( \delta {X}^T \mathbf{Q} \delta{X} + \delta{u}^T \mathbf{R} \delta{u} \right) dt ,
\end{equation}
with $\delta {X} = {X} - {X}_d$, $\delta{u} = {u} - {u}_0$. The weighting matrices are:
\begin{align*}
\mathbf{Q} &= \text{diag}(200, 200, 50, 5, 5, 2, 10, 10, 1, 1, 1, 1), \\
\mathbf{R} &= \text{diag}(4, 4, 2, 2),
\end{align*}
emphasizing position tracking over attitude and penalizing servo angles less than rotor speeds. The optimal gain $\mathbf{K} = \mathbf{R}^{-1} \mathbf{B}^T \mathbf{P}$ is computed by solving the Algebraic Riccati Equation (26).
\begin{equation}
	\mathbf{A}^T \mathbf{P} + \mathbf{P} \mathbf{A} - \mathbf{P} \mathbf{B} \mathbf{R}^{-1} \mathbf{B}^T \mathbf{P} + \mathbf{Q} = 0 .
\end{equation}

\section{Simulation Results}

\subsection{Simulation with Torque Inputs}
The simulation parameters for the torque input are detailed in Table \ref{tab:sim_params}.

\begin{figure}[htbp]
\centering
\resizebox{0.5\textwidth}{!}{\includegraphics[trim=0 150 0 150,clip]{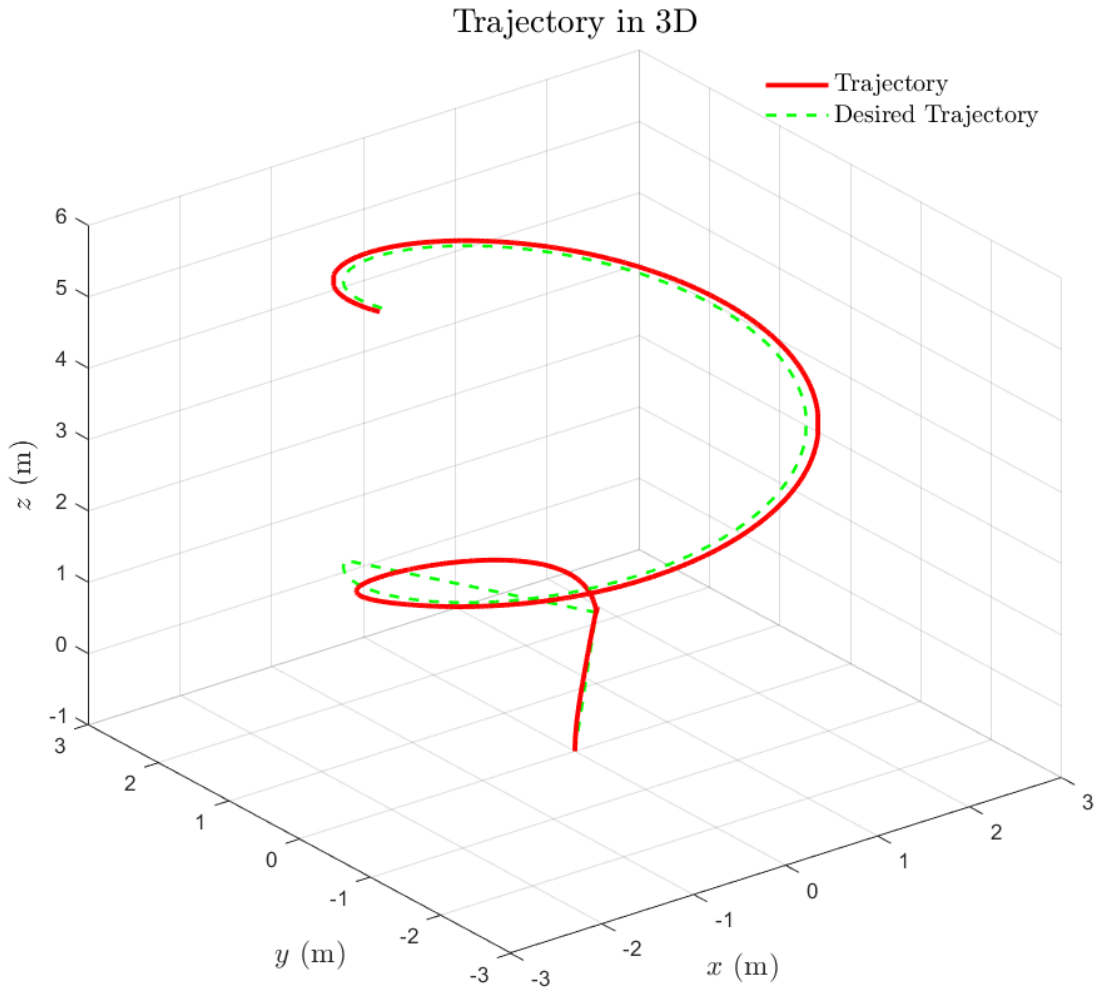}}
\caption{3D trajectory of the UAV with torque inputs as control signals}
\label{tau_3D}
\end{figure}

\begin{figure}[htbp]
\centering
\resizebox{0.5\textwidth}{!}{\includegraphics[trim=0 150 0 150,clip]{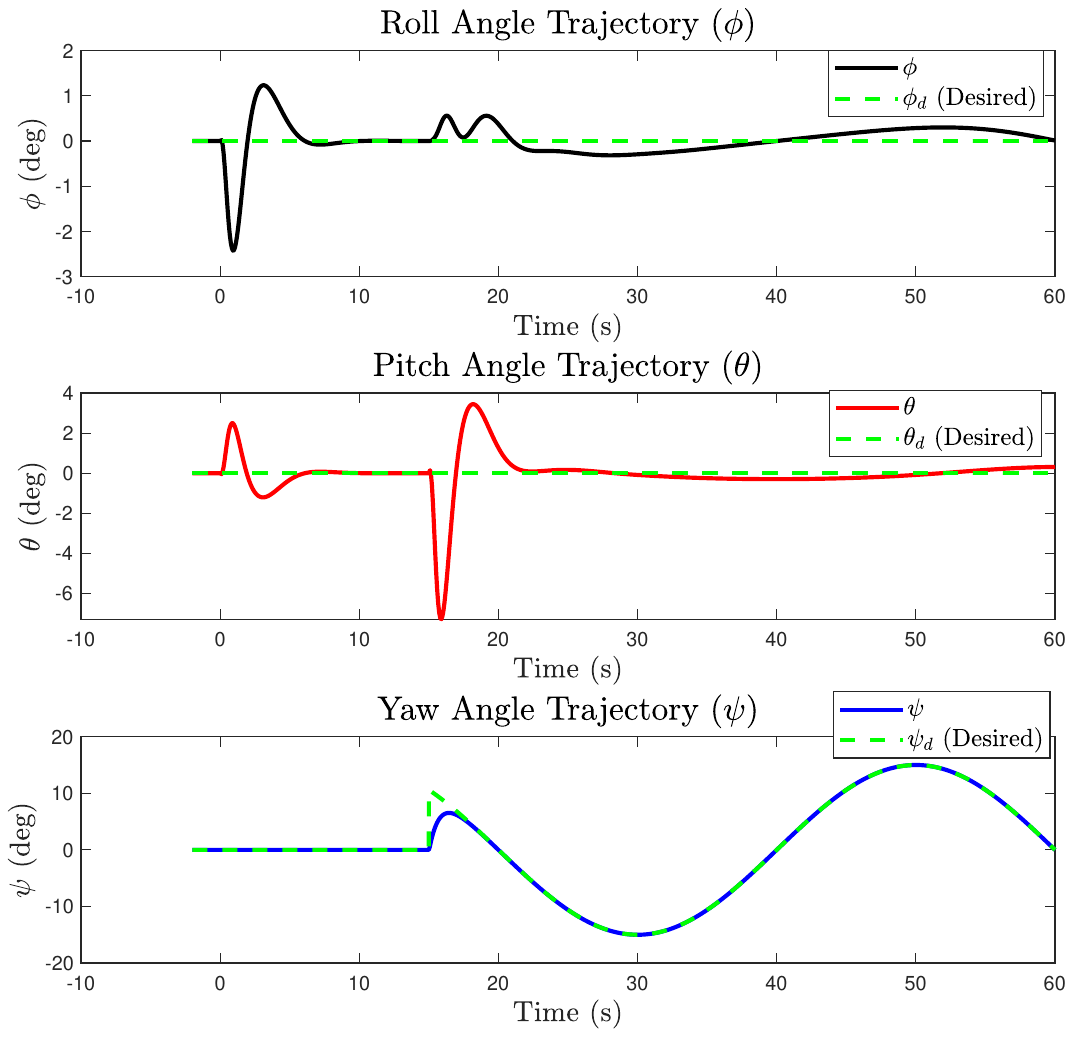}}
\caption{Roll, pitch, and yaw angles of the UAV with torque inputs as control signals}
\label{tau_rpy}
\end{figure}

\begin{figure}[htbp]
\centering
\resizebox{0.5\textwidth}{!}{\includegraphics[trim=0 150 0 140,clip]{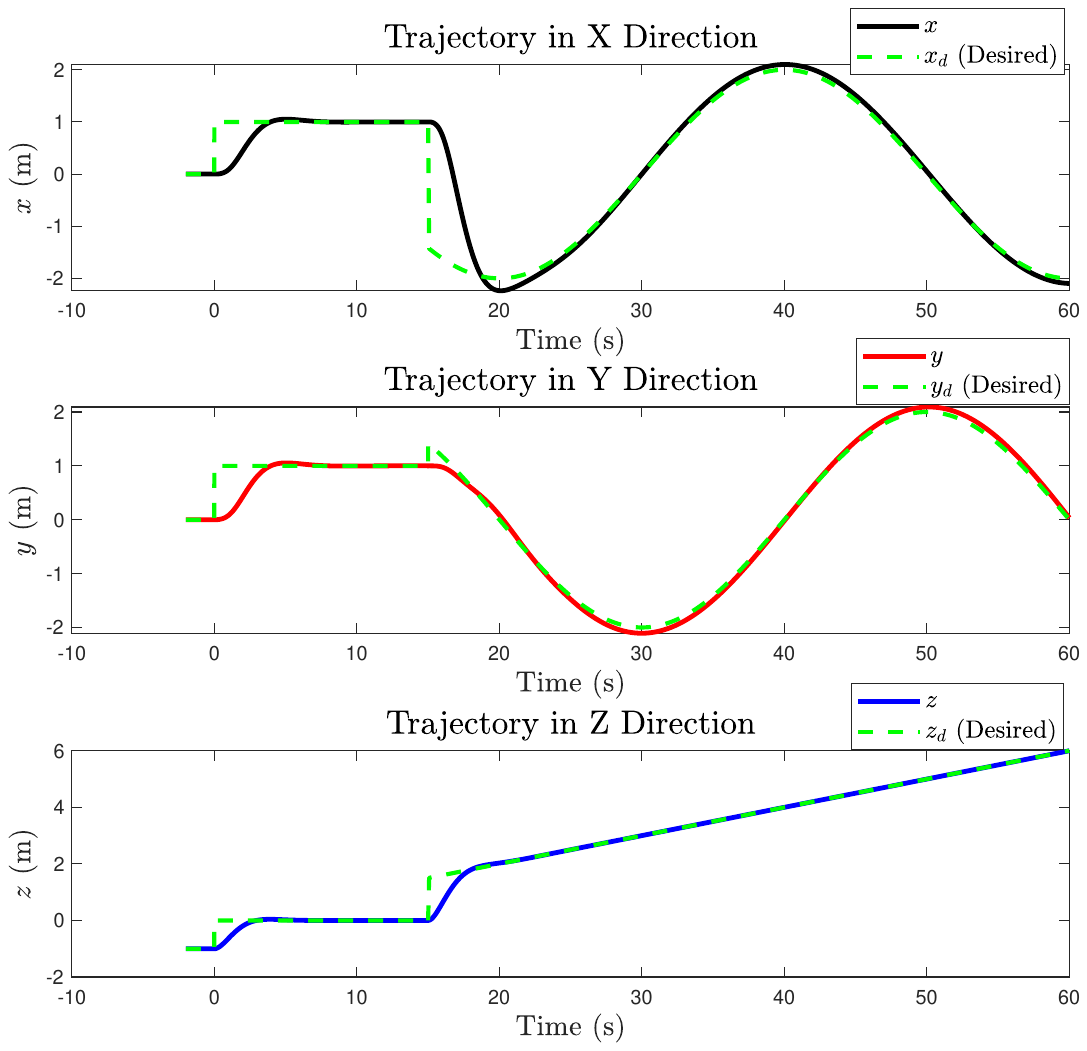}}
\caption{Position of the UAV with torque inputs as control signals}
\label{tau_xyz}
\end{figure}

\begin{figure}[htbp]
\centering
\resizebox{0.5\textwidth}{!}{\includegraphics[trim=0 150 0 150,clip]{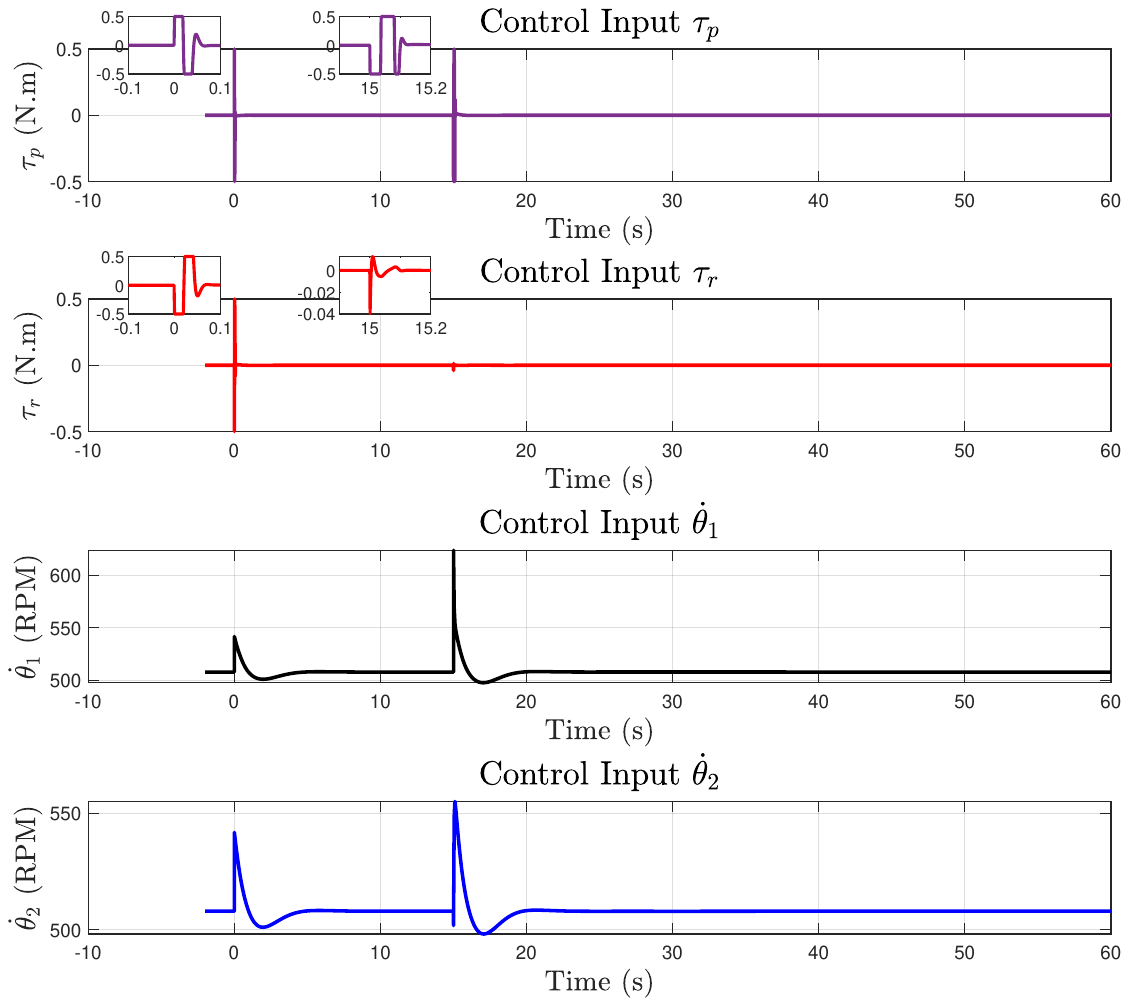}}
\caption{Control signals of the UAV with torque inputs as control signals}
\label{tau_u}
\end{figure}

\begin{table}[htbp]
	\centering
	\caption{Simulation parameters for the torque input scenario}
	\begin{center}
		\begin{tabular}{|c|c|c|}
			\hline
			\textbf{Parameter} & \textbf{Value} & \textbf{Unit} \\
			\hline
			Solver & \texttt{ode45} & - \\
			Time step ($\Delta t$) & 0.001 & s \\
			Simulation duration ($t$) & [-2, 60] & s \\
			\hline
		\end{tabular}
	\end{center}
	\label{tab:sim_params}
\end{table}
Figures \ref{tau_3D}, \ref{tau_rpy}, and \ref{tau_xyz} show the three-dimensional trajectory, body attitudes, and position of the UAV in this simulation. The results confirm that the UAV is able to take off, hold steady hover, and subsequently follow the circular trajectory with rising altitude. Tracking errors are small, although there is a slight lag in the x-direction when in circular motion. Figure \ref{tau_u} shows control signals, where exhibit torque inputs have reached saturation that has been applied in simulation. The abrupt change at 15 seconds corresponds to the switching between hover mode and the ascending circular trajectory.

\subsection{Simulation with Sensor Noise and EKF}
\begin{figure}[htbp]
\centering
\resizebox{0.5\textwidth}{!}{\includegraphics[trim=0 150 0 130,clip]{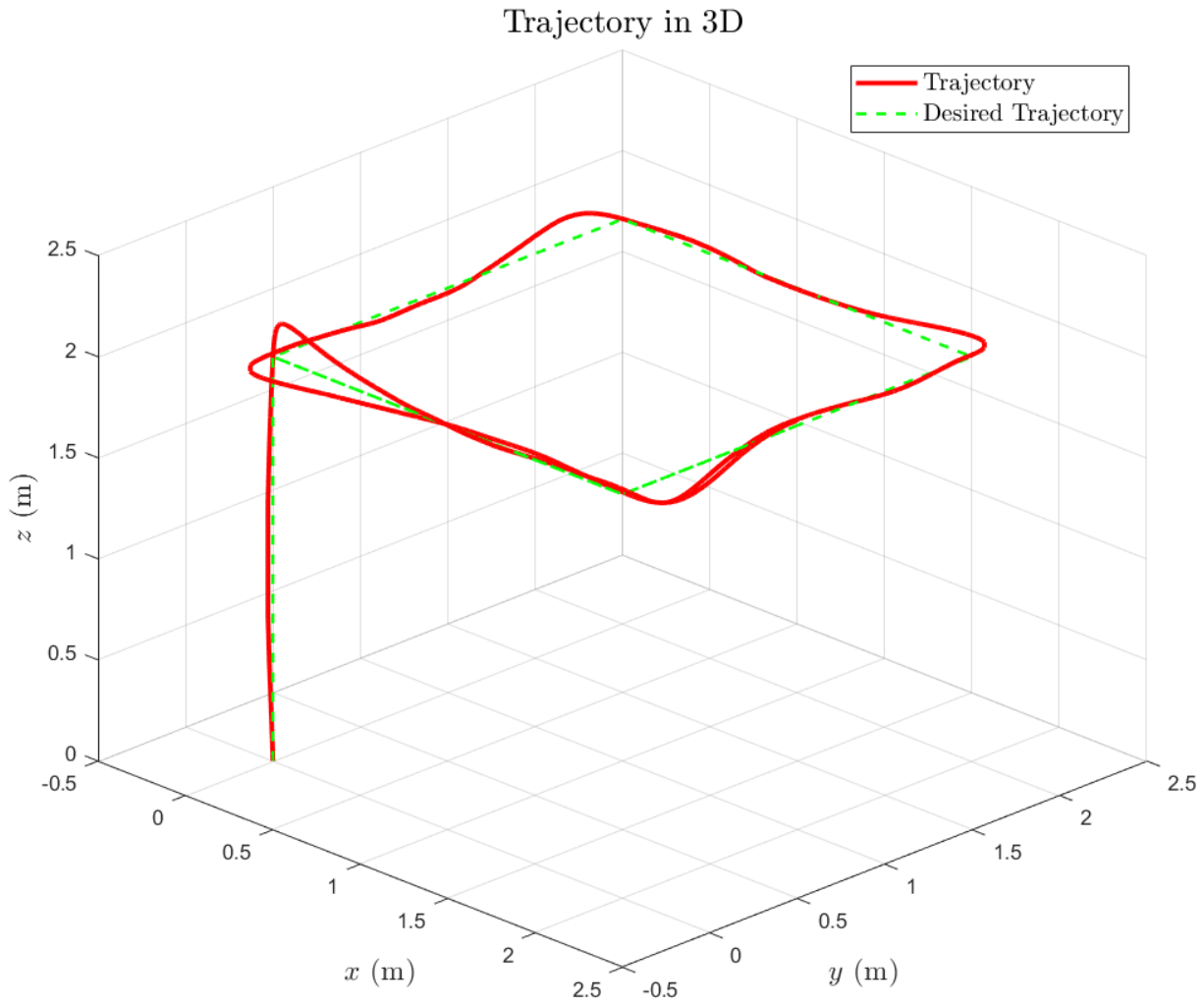}}
\caption{3D trajectory of the UAV with EKF}
\label{kalman_3D}
\end{figure}

\begin{figure}[htbp]
\centering
\resizebox{0.5\textwidth}{!}{\includegraphics[trim=0 150 0 150,clip]{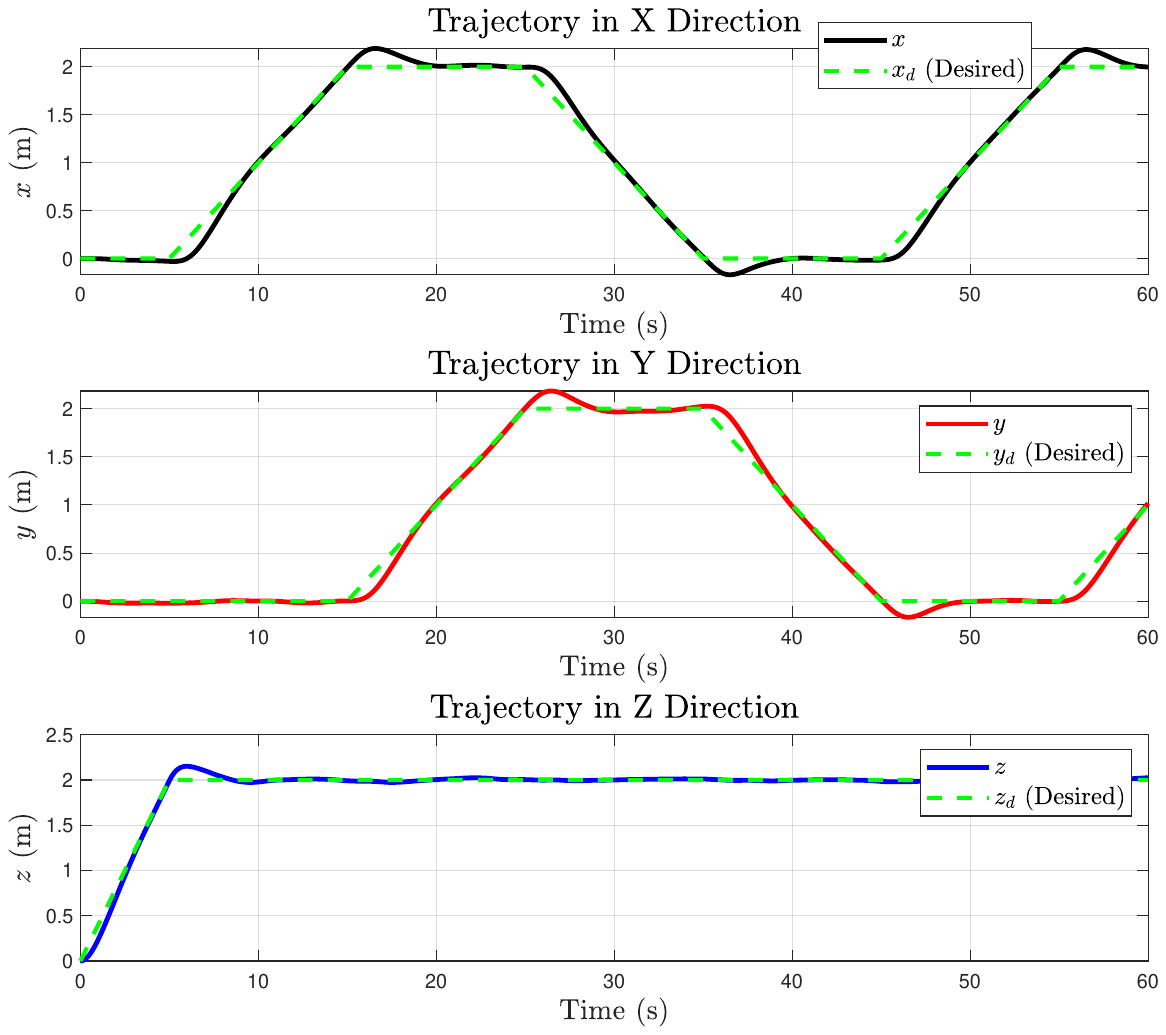}}
\caption{Position of the UAV with EKF}
\label{kalman_xyz}
\end{figure}

\begin{figure}[htbp]
\centering
\resizebox{0.5\textwidth}{!}{\includegraphics[trim=0 150 0 150,clip]{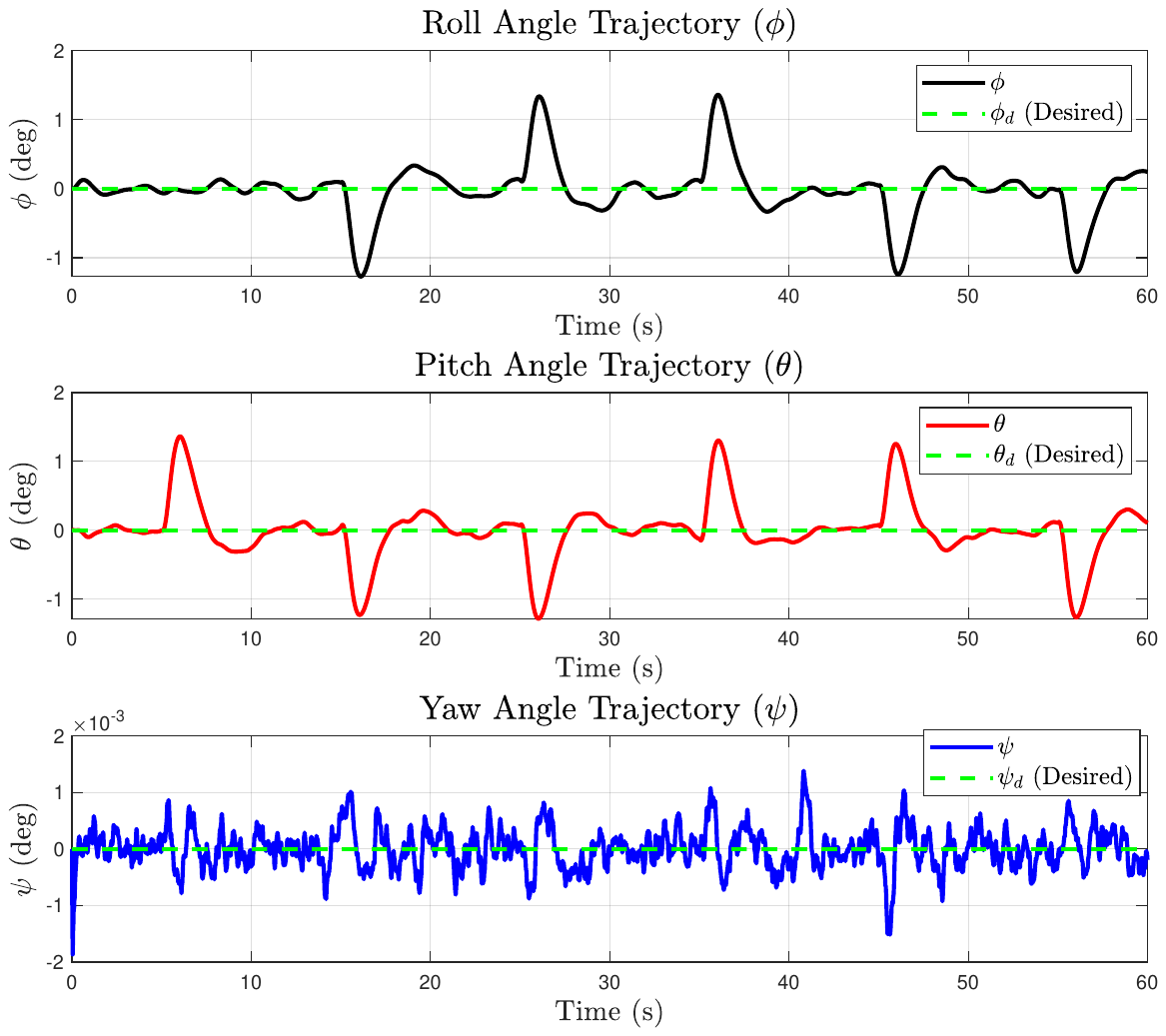}}
\caption{Roll, pitch, and yaw angles of the UAV with EKF}
\label{kalman_rpy}
\end{figure}

\begin{figure}[htbp]
\centering
\resizebox{0.5\textwidth}{!}{\includegraphics[trim=0 160 0 150,clip]{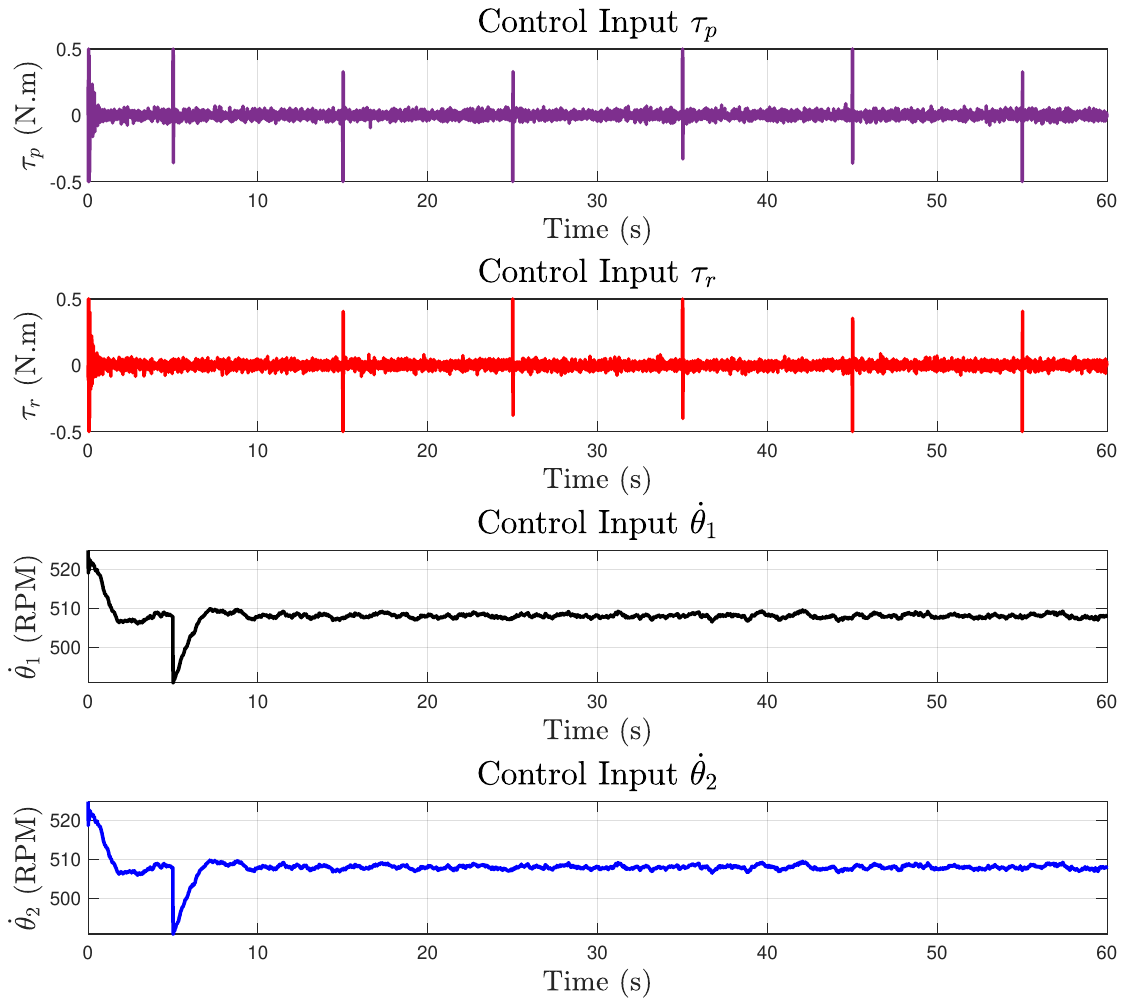}}
\caption{Control signals of the UAV with EKF}
\label{kalman_u}
\end{figure}

\begin{figure}[htbp]
\centering
\resizebox{0.5\textwidth}{!}{\includegraphics[trim=0 170 0 150,clip]{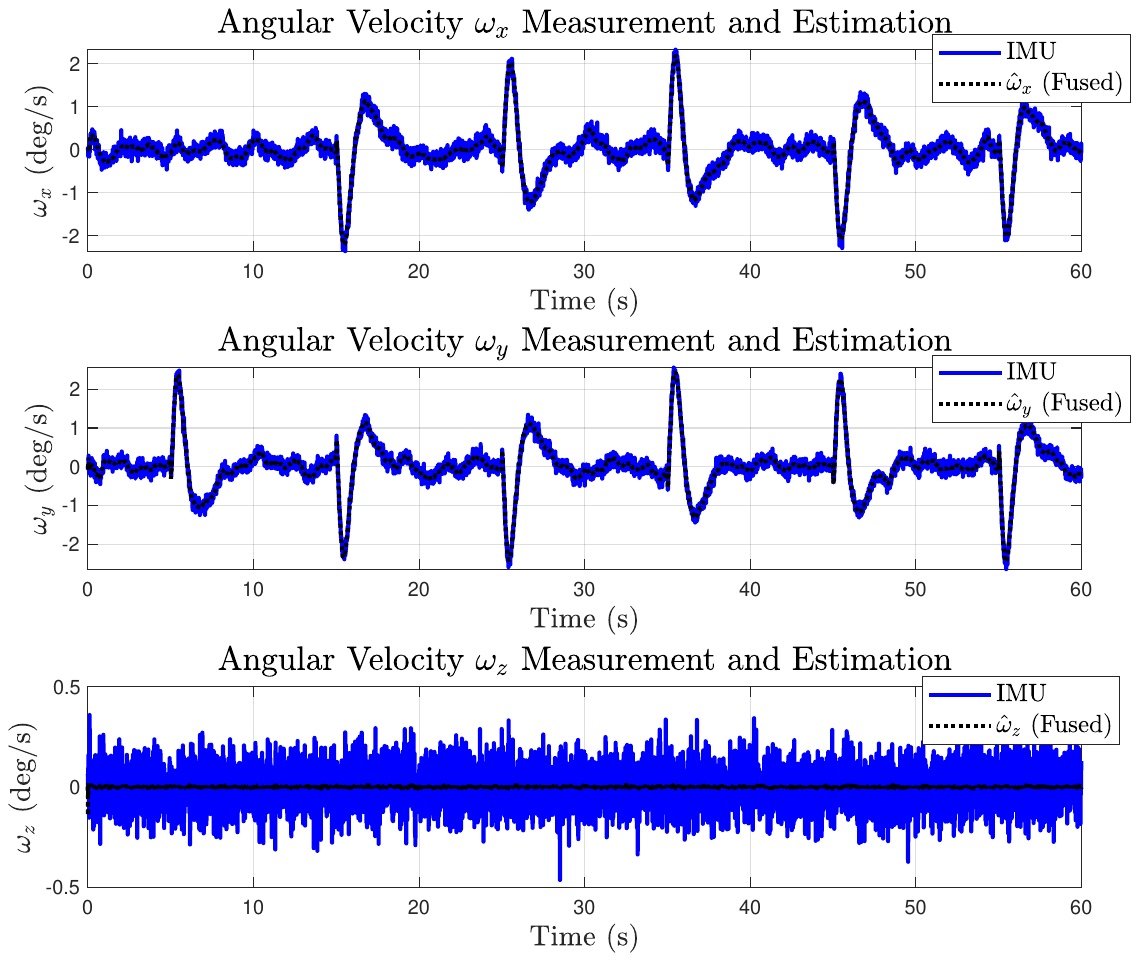}}
\caption{Filtering IMU sensor data using the EKF}
\label{kalman_w}
\end{figure}

\begin{figure}[htbp]
\centering
\resizebox{0.5\textwidth}{!}{\includegraphics[trim=0 170 0 150,clip]{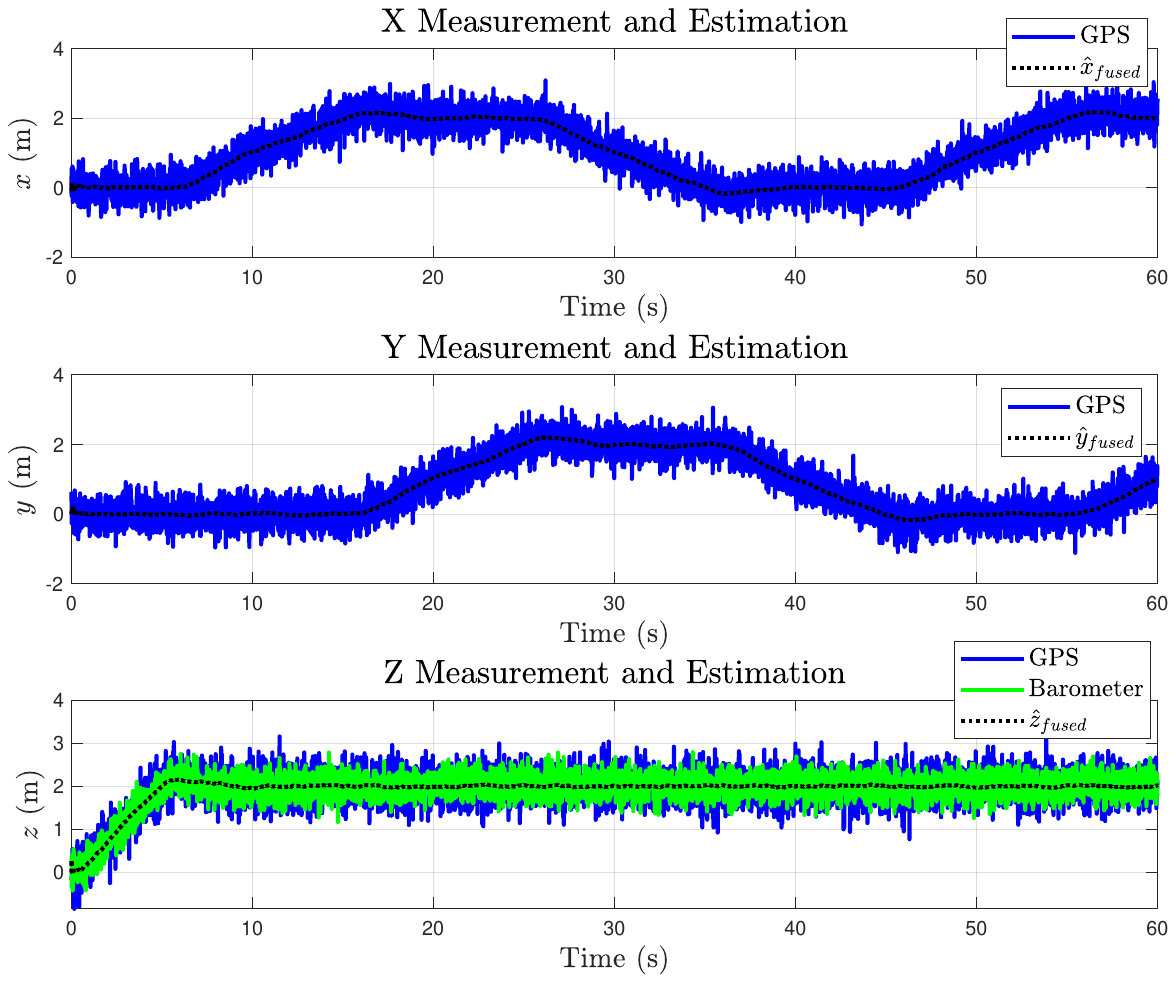}}
\caption{Fusion of GPS and barometer data using the EKF}
\label{kalman_x}
\end{figure}

A constant altitude square reference trajectory was used in this simulation. The results completely demonstrate the performance of the Extended Kalman Filter for estimating system states.
The 3D path of the UAV (Figure \ref{kalman_3D}) and the time positions (Figure \ref{kalman_xyz}) confirm that the filtered estimates enable the controller to achieve excellent performance, causing the UAV to follow the reference trajectory with excellent precision.
The roll, pitch, and yaw Euler angles (Figure \ref{kalman_rpy}) also exhibit good stability in the rotational dynamics of the UAV.
Control signals used (Figure \ref{kalman_u}) are optimal but oscillatory due to sensor noise. Overall, the results show proper controller operation along with the filter. Raw IMU sensor readings (Figure \ref{kalman_w}) and GPS/barometer readings (Figure \ref{kalman_x}) show that the EKF can successfully reduce measurement noise by a significant amount and provide state estimates with high accuracy. Sensor data fusion of these sensors leads to a significant improvement in position and angle estimation.


\section{Discussion and Conclusion}
The study challenged the inherent underactuation of coaxial UAVs using the innovative two-degrees-of-freedom pendulum thrust vectoring mechanism. The key contribution is a clear-cut Lagrangian dynamic model that for the first time fully captures the inertial effects of all the actuated units and provides a firm basis for control synthesis. Based on this model, an LQR controller was simulated and verified by high-fidelity simulations, accomplishing complex maneuvers such as takeoff, hover, and 3D trajectory following. An Extended Kalman Filter (EKF) extension was also applied, which further enhanced performance by allowing robust state estimation with realistic sensor noise.

The active thrust vectoring provided by the proposed 2-DOF pendulum mechanism resolves the inherent underactuation in a mechanically elegant design and offers full roll and pitch control. Further, our EKF plays an important role in making the LQR controller reliable by providing multi-sensor data fusion-based accurate state estimates which are robust against real-world noise and dynamics.

While these simulations confirm the framework, experimental verification is the primary limitation. Physical realization of the system is therefore the subsequent step. This involves the development of a model and performing flight tests to examine the theoretical models against actual conditions, including aerodynamic disturbances and hardware constraints. Further research will also examine modeling enhancement of the aerodynamic model and use of advanced nonlinear controllers as one way of increasing the platform's robustness and agility in real-world applications.


\begin{thebibliography}{00}
\bibitem{b1} Giernacki, W., Gośliński, J., Goślińska, J., Espinoza-Fraire, T.,  Rao, J. (2021). Mathematical modeling of the coaxial quadrotor dynamics for its attitude and altitude control. Energies, 14(5), 1232.

\bibitem{b2} Chen, L., Xiao, J., Zheng, Y., Alagappan, N. A., Feroskhan, M. (2024). Design, modeling, and control of a coaxial drone. IEEE Transactions on Robotics, 40, 1650-1663.

\bibitem{b3} Dominguez, V. H., Garcia-Salazar, O., Amezquita-Brooks, L., Reyes-Osorio, L. A., Santana-Delgado, C., Rojo-Rodriguez, E. G. (2022). Micro coaxial drone: flight dynamics, simulation and ground testing. Aerospace, 9(5), 245.

\bibitem{b4} HajiAbedini, M., Zargarbashi, F., Talaeizadeh, A., Pishkenari, H. N., Alasty, A. (2021, November). Design and implementation of a novel over-actuated quadrotor with variable dihedral angle. In 2021 9th RSI international conference on robotics and mechatronics (ICRoM) (pp. 391-398). IEEE.

\bibitem{b5} Rashad, R., Goerres, J., Aarts, R., Engelen, J. B., Stramigioli, S. (2020). Fully actuated multirotor UAVs: A literature review. IEEE Robotics and Automation Magazine, 27(3), 97-107.

\bibitem{b6} North, A. (2023). Modeling, Control, and Hardware Development of a Thrust-Vector Coaxial UAV.

\bibitem{b7}  Glida, H. E., Sentouh, C., Rath, J. J. (2023). Optimal model-free finite-time control based on terminal sliding mode for a coaxial rotor. Drones, 7(12), 706.

\bibitem{b8} Wei, Y., Chen, H., Li, K., Deng, H., Li, D. (2019). Research on the control algorithm of coaxial rotor aircraft based on sliding mode and PID. Electronics, 8(12), 1428.

\bibitem{b9} Bernardes, E., Boyer, F., Viollet, S. (2023). Modelling, control and simulation of a single rotor UAV with swashplateless torque modulation. Aerospace Science and Technology, 140, 108433.

\bibitem{b10} Kim, D. K., Yoon, B. I., Song, Y. H., Song, J. (2018). An experimental study on the hover performance characteristics of the coaxial propellers configuration for the drone.

\bibitem{b11} Yoon, S., Chan, W. M., Pulliam, T. H. (2017). Computations of torque-balanced coaxial rotor flows. In 55th AIAA aerospace sciences meeting (p. 0052).

\bibitem{b12} Panjwani, B., Quinsard, C., Przemysław, D. G., Furseth, J. (2020). Virtual Modelling and Testing of the Single and Contra-Rotating Co-Axial Propeller. Drones, 4(3), 42.

\bibitem{b13} Li, K., Wei, Y., Wang, C., Deng, H. (2019). Longitudinal attitude control decoupling algorithm based on the fuzzy sliding mode of a coaxial-rotor UAV. Electronics, 8(1), 107.

\bibitem{b14} Wang, S., Li, K., Chen, H., Deng, H. (2019, October). Attitude control of small coaxial twin-rotor aircraft. In 2019 IEEE International Conference on Unmanned Systems (ICUS) (pp. 558-562). IEEE.

\bibitem{b15} Wang, Y., Wang, Z., Wang, G., Chen, L. (2024). Modeling and Control of A Coaxial Pendulum Drone. IEEE Transactions on Intelligent Vehicles.

\end{thebibliography}
\end{document}